\documentclass{article} 
\usepackage[final]{colm2026_conference}

\usepackage{microtype}
\usepackage{hyperref}
\usepackage{url}
\usepackage{booktabs}
\usepackage{amsmath}
\usepackage{graphicx}
\usepackage{pifont}

\usepackage{lineno}

\definecolor{darkblue}{rgb}{0, 0, 0.5}
\hypersetup{colorlinks=true, citecolor=darkblue, linkcolor=darkblue, urlcolor=darkblue}

\title{Synchronized Logit Steering: Real-world Steganography}

\author{%
  Andrew Rufail, Aadi Dash, Onir Narahari, Ethan Mui, \\
  \textbf{Mahi Gajare, Prakhar Tiwari, Shrija Makapothula, Nick Cui} \\
  Algoverse AI \\
  \texttt{andrew.rufail@gmail.com} 
}

\newcommand{\cmark}{\ding{51}}
\newcommand{\xmark}{\ding{55}}

\begin{document}

\ifcolmsubmission
\linenumbers
\fi

\maketitle

\begin{abstract}
  Steganography in large language models offers a way to embed hidden messages within natural-sounding text. Existing token and logit-level methods typically require the sender and receiver to share an identical prompt context, which is rarely guaranteed in production pipelines that use retrieval-augmented generation or proprietary system instructions. We introduce Synchronized Logit Steering (SLS), a deterministic steganographic scheme that eliminates this dependency by deriving a proxy prompt from the generated output itself, allowing both parties to reconstruct the same logit distribution without access to the original prompt. SLS encodes payload values as token ranks within high-entropy regions of the proxy prompt distribution, and we extend the scheme with periodic recurrence and payload bursts to scale information density. Across ShareGPT, GSM8K, and SWE-bench Verified, we show that the KL divergence between the true and proxy prompt distributions falls below 0.5 nats once the synchronization window reaches 40 tokens, and SLS encoding does not meaningfully disrupt this convergence relative to greedy generation. We also find that the periodic-burst variant achieves 0.20 bits per token, or roughly $10\times$ the capacity of single-payload encoding. Kolmogorov-Smirnov tests further confirm that SLS outputs are statistically difficult to distinguish from greedy generations, demonstrating that covert, prompt-agnostic communication through LLMs is both practical and stealthy.
\end{abstract}

\section{Introduction}

Steganography is the practice of concealing information such that the presence of the hidden message is obscured within seemingly ordinary content, making detection and tampering significantly more difficult \citep{bai_semantic_2024,  channalli_steganography_2009, perry_robust_2025}. Unlike cryptography, which focuses on rendering messages unreadable to unauthorized readers, steganography aims to hide the very existence of the message itself.

To frame the concept, we draw upon the "prisoners' problem" \citep{simmons_the_1984, wu_generative_2024}, which demonstrates the principle of steganography. In this scenario, two prisoners, Alice and Bob, must communicate secretly under the warden's eye, Eve, who may attempt to censor or manipulate their communication. By embedding secret messages within innocent-looking content, Alice attempts to evade Eve’s inspection, even in the presence of tampering or deception.

Large language models (LLMs) provide a natural setting for this problem. They generate text based on a probability distribution over possible next tokens. At each generation step, multiple high-probability tokens can produce equally natural continuations, resulting in inherent variability in model outputs. This means that different valid words can be chosen at each step, allowing information to be encoded in these choices while still producing natural-sounding text.

This study investigates the steganographic potential of contemporary LLMs. Specifically, we examine whether hidden messages can be embedded directly within the model’s token generation process while remaining indistinguishable from ordinary text. For successful decoding, the encoding must be accurate and consistent across generations. Unlike previous approaches that rely on prompt engineering or surface-level manipulation, our method embeds information through token selection based on the model’s probability distribution. Despite this potential, many existing steganographic techniques remain fragile or easy to manipulate. For example, simple transformations, such as paraphrasing or reformatting, can destroy hidden messages while preserving the apparent meaning of the text.

 Existing approaches to steganography in language models often rely on prompt engineering or surface-level manipulation of generated text. These methods are not deterministic, as the way the hidden message is encoded depends on how the model responds to a prompt rather than a fixed rule. As a result, the encoded information can be inconsistent across generations and difficult to reliably decode. This limits their effectiveness for consistent message recovery and shows the need for a method that controls the token generation process directly.

In this work, we introduce a deterministic steganographic method that encodes information directly within the token generation process of LLMs. Instead of relying on prompt behavior, the hidden message is embedded through a fixed encoding rule based on the model’s next-token probabilities. We show that the method enables consistent decoding while preserving the fluency of the generated text. These results demonstrate that direct control over token generation enables practical covert communication in LLMs. Future work may explore techniques to safeguard or disrupt such encoding, depending on the intended application.

\section{Related Works} 

Large language models (LLMs) can be fine-tuned or trained to encode and decode hidden information in text \citep{mathew_hidden_2024,li2023stegolm}. However, existing approaches fall into three broad classes, each with distinct limitations that motivate our approach.

A first class of methods trains dedicated encoder-decoder models to embed and recover hidden messages. \citet{bieniasz_large_2024} train two agents, an encoder and a decoder, through a feedback loop to embed and recover hidden messages in a black-box environment. Similarly, \citet{mathew_hidden_2024} and \citet{karpov_steganographic_2023} explore encoder–decoder frameworks in which reinforcement learning or structured prompting enables secret communication between models. These studies demonstrate that LLMs can develop emergent communication strategies, though they are typically limited to short payloads or controlled experimental settings.

A second class relies on prompt engineering or surface-level manipulation of generated text to encode hidden information. For example, \citet{zhou2025autostega} use an agent-driven framework to generate prompt-based strategies for embedding hidden information in text, while \citet{raz2026canaries} encode identifiers using linguistic and formatting-based transformations within otherwise natural-looking documents. However, because the encoding relies on the exact generated output, these methods are not deterministic and can be disrupted by minor modifications to the text.

A third class of methods encodes hidden information directly through token- or logit-level decisions. Since LLMs generate text by assigning probabilities to possible next tokens, the sender can map parts of a secret message to token choices within the model’s high-probability set. Similar probability-based approaches in \citet{wu_generative_2024} and \citet{perry_robust_2025} encode hidden information through controlled sampling from the model’s next-token distribution. However, these methods often require Alice and Bob to share the same model, decoding rule, and generation context. As a result, differences in context can change the probability distribution and cause decoding to fail.

These limitations motivate a method that preserves the systematic structure of token-level encoding while reducing dependence on a hidden or mismatched prompt context. Table~\ref{tab:method-comparison} summarizes these limitations alongside SLS, which addresses all three. Our method achieves this by using synchronized logit steering, where Alice and Bob decode from a shared proxy prompt derived from the generated output itself rather than relying on access to the entire original prompt.

\begin{table}[h]
\centering
\caption{Comparison of steganographic approaches across key properties. Method categories follow the taxonomy in Section 2.}
\label{tab:method-comparison}
\begin{tabular}{lcccc}
\toprule
\textbf{Method} & \textbf{Deterministic} & \textbf{No Shared} & \textbf{Robust to} & \textbf{High} \\
 & \textbf{Decoding} & \textbf{Context} & \textbf{Text Edits} & \textbf{Capacity} \\
\midrule
Fine-tuned Encoder-Decoder & \cmark & \cmark & \cmark & \xmark \\
Prompt Engineering         & \xmark & \cmark & \xmark & \cmark \\
Token/Logit Sampling       & \cmark & \xmark & \cmark & \cmark \\
\textbf{SLS}        & \cmark & \cmark & \cmark & \cmark \\
\bottomrule
\end{tabular}
\end{table}

\section{Methodology}

We introduce Synchronized Logit Steering (SLS), a deterministic steganographic scheme designed for environments where the sender (Alice) and receiver (Bob) share a language model $\mathcal{M}$ but do not know each other and lack a shared input prompt.

\subsection{Problem Definition}
Existing linguistic steganography techniques that leverage logit ranks typically assume that both parties have access to a shared context $\mathcal{C}$ to compute identical logit distributions $P(y_t|y_{<t}, \mathcal{C})$ \citep{jiang2025high, ziegler2019neural, ding2023discop}. However, in production LLM pipelines, particularly those utilizing Retrieval-Augmented Generation \citep{lewis2020retrieval} or hidden system-level instructions \citep{jaech2024openai}, the full context $\mathcal{C}$ is often non-deterministic or proprietary. 

If Alice attempts to encode a message $m$ using the distribution $P(\cdot|Prompt_{ALICE})$, Bob will fail to decode it if his available context $Prompt_{Bob} \neq Prompt_{Alice}$. To resolve this, we propose using a Proxy prompt derived purely from the generated output tokens. 

\subsection{Proxy Prompt}

Let $y = {y_1, y_2, ..., y_N}$ be the sequence of tokens generated by the model. We define a synchronization window of length $n$. The first n tokens, $y_{1:n}$, are generated normally through the primary prompt $P_{Alice}$. For subsequent tokens, we defined a proxy prompt $P'=y_{1:n}$.

Encoding a payload of value $v \in {0,...,K-1}$ occurs at index $n+1$. The encoder computes the logit distribution on the proxy prompt with a temperature of zero (no sampling).
$$\pi_{t} = \mathcal{M}(y_{<t} = P')$$
The tokens are sorted by logarithmic probability in descending order. The steganographic token $y_{n+1}$ is selected such that rank$(y_{n+1})=v$. Since Bob receives the same text $y$, he can reconstruct $P'$ and  $\pi_{t}$ to extract $v$ by reversing the process.
$$v = \text{rank}_{\pi_t}(y_{n+1})$$

\subsection{Preventing detection}

To maintain statistical indistinguishability from the normal LLM language, SLS only executes encoding when the local entropy of the proxy prompt exceeds a threshold $\tau$ (set to $\tau = 2.5$ in our experiments). Furthermore, the window size $n$ has to be optimized to minimize the Kullback–Leibler (KL) divergence \citep{kullback1951information} between the original prompt and proxy prompt distributions.
$$\min_n D_{KL} \left( P(\cdot | y_{<t}, P) \parallel P(\cdot | y_{<t}, P') \right)$$
This ensures that using the output prefix as a proxy prompt does not significantly alter the trajectory of the original prompt logic. In practice, we have found that across models and tasks, $n=40$ is an acceptable value to use.

\subsection{Payload Capacity}
The information density, called Bits Per Token (BPT), is calculated as $BPT = \frac{k}{N}$, where a single payload of bit width $k=log_2K$ is embedded in an output sequence of length $N$.

To scale capacity, we extend SLS using two strategies:

\begin{enumerate}
\item \textbf{Periodic:} The encoding process is repeated every $n$ steps, treating all the preceding output tokens as the updated proxy prompt.

\item \textbf{Bursted:} We utilize the proxy prompt to predict a sequence of $a$ consecutive steganographic tokens, given the same entropy constraint.
\end{enumerate}

Assuming all tokens meet the entropy requirements, the generalized capacity of this periodic burst configuration is defined as:
$$BPT = \frac{a \cdot \log_2 K}{n + a}$$
This formulation allows for a tunable trade-off between the detectability of the text and the capacity of the payload.

\section{Results}
We evaluate the performance of SLS using \texttt{gemma-2-2b} \citep{gemma2024} across three diverse datasets: ShareGPT\footnote{The ShareGPT dataset is collected from \url{https://sharegpt.com}.} for general conversations, GSM8K \citep{cobbe2021gsm8k} for mathematical reasoning, and SWE-bench Verified \citep{jimenez2023swebench} for software engineering. All experiments use a synchronization window of $n = 40$, entropy threshold $\tau = 2.5$, rank depth $K = 16$ (4 bits per token), and burst size $b = 5$. Our analyses focus on the convergence of the proxy prompt, the resulting payload capacity, and the statistical stealth of the generated text.

The efficacy of SLS relies on the proxy prompt $P'$ accurately approximating the original distribution of $P$. We measured the Kullback-Leibler (KL) divergence between these two distributions as a function of the synchronization window $n$ across all three evaluation datasets and encoding variants. 

\begin{figure}[h]
    \centering
    \includegraphics[width=\columnwidth]{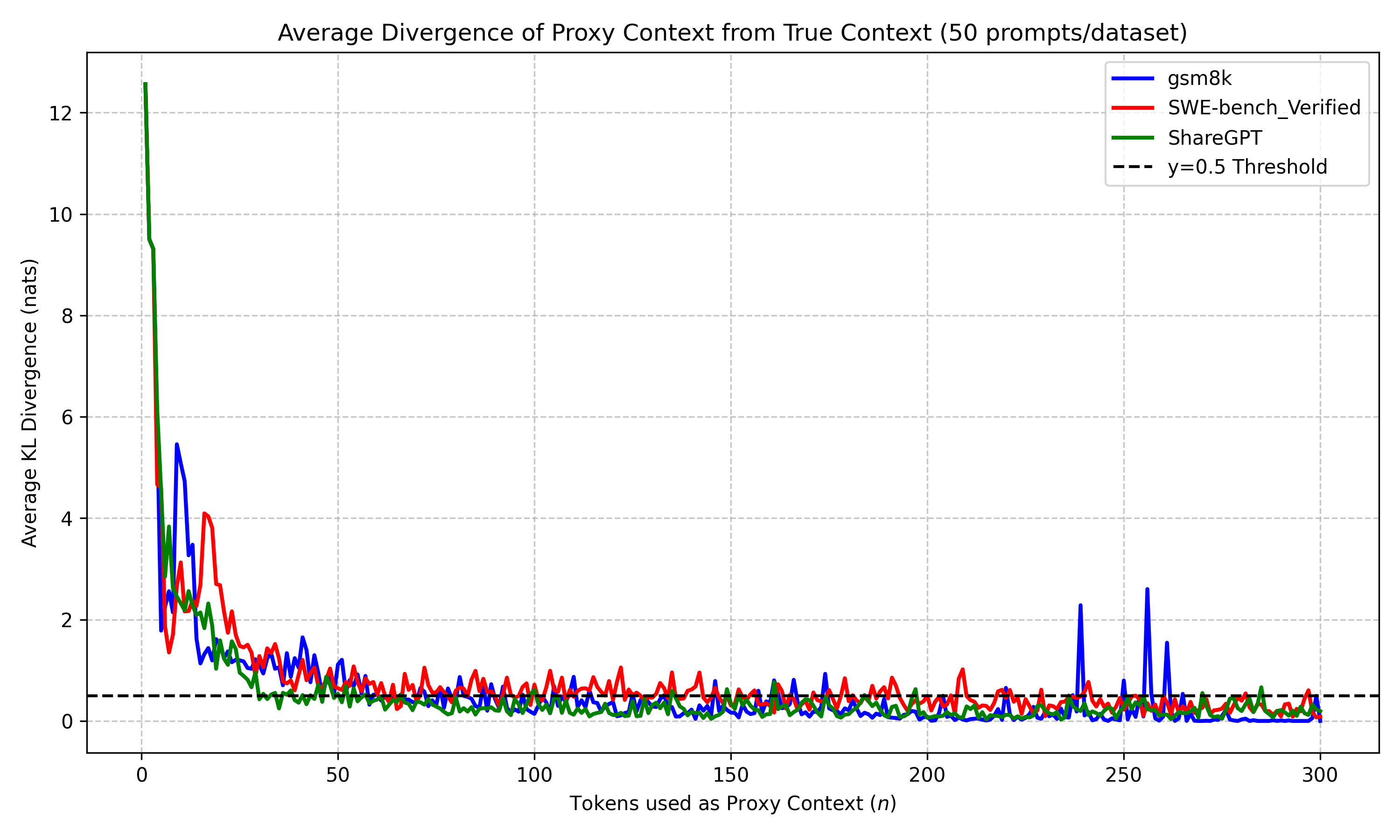}
    \caption{Average KL divergence between the true prompt distribution and the proxy prompt distribution, averaged over 50 prompts per dataset. The divergence approaches 0.5 nats at $n = 40$ and stabilizes fully by $n \approx 50$.}
    \label{fig:kl_main}
\end{figure}

\begin{figure}[h]
    \centering
    \includegraphics[width=\columnwidth]{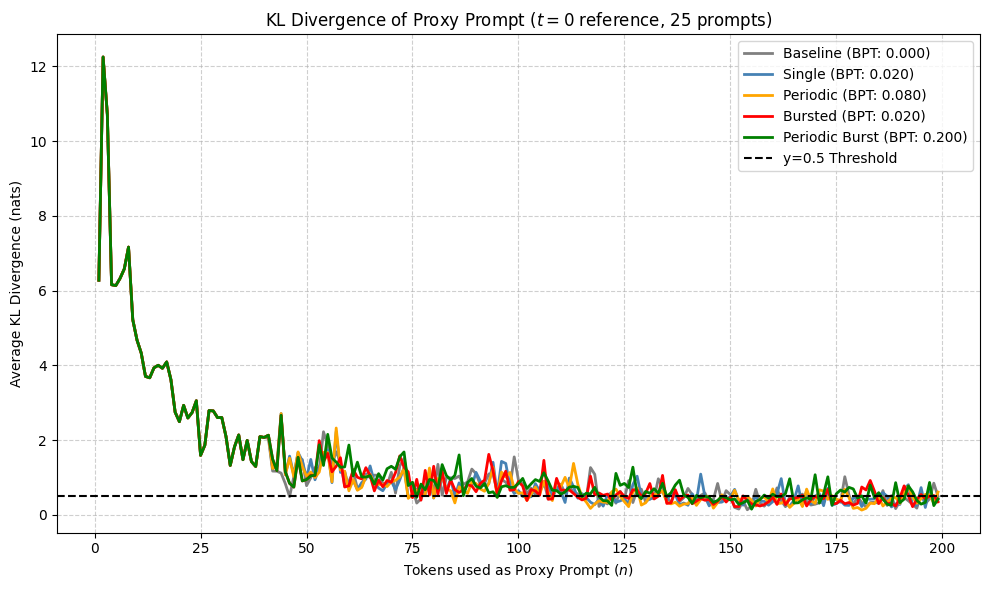}
    \caption{Average KL divergence between the true prompt distribution and the proxy prompt distribution across SLS encoding variants, averaged over 25 ShareGPT prompts. All SLS variants closely track the greedy baseline trajectory, confirming that encoding does not meaningfully shift the proxy prompt distribution relative to normal generation.}
    \label{fig:kl_methods}
\end{figure}

Across all datasets, the KL divergence is initially high but decays rapidly as more tokens are added to the proxy prompt. As shown in Figure \ref{fig:kl_main}, the divergence consistently falls below our target threshold of 0.5 nats once $n \geq 40$, justifying our choice of synchronization window across the three datasets tested. Figure \ref{fig:kl_methods} shows the KL divergences across SLS encoding strategies averaged over 25 ShareGPT prompts. Notably, all SLS strategies closely track the baseline throughout the entire output length, confirming that steganographic encoding does not meaningfully disrupt the proxy prompt distribution relative to greedy generation.

Furthermore, we compared the information density using BPT according to the strategies described in the Methodology section to increase capacity: Bursted, Periodic, and Periodic Burst. Table~\ref{tab:payload_data} summarizes the BPT and GSM8K accuracy results across all four encoding strategies.

\begin{itemize}
    \item \textbf{Semantic convergence}. As shown in Figure \ref{fig:kl_methods}, all SLS variants track the greedy baseline KL divergence closely across the full output length.
    \item \textbf{Information Density}. Periodic Burst encoding provides the highest encoding capability, having $10\times$ more capacity compared to single payload and bursted encoding, and $2.5\times$ capacity compared to periodic encoding (Table \ref{tab:payload_data}). This would make it an ideal choice for real steganographic systems.
\end{itemize}

\begin{table}[h]
\caption{Payload capacity over a 200-token output.}
\label{tab:payload_data}
\vskip 0.15in
\begin{center}
\begin{small}
\begin{sc}
\begin{tabular}{lccc}
\toprule
Method & BPT & Payload &GSM8K Acc\\
\midrule
None (baseline) & - & - & 0.73 \\
Single & 0.0200 & 4.0 & 0.66 \\
Bursted & 0.0200 & 4.0 & 0.60\\
Periodic & 0.0800 & 16.0 & 0.68\\
PERIODIC BURST & \textbf{0.2000} & \textbf{40.0} & 0.57 \\
\bottomrule
\end{tabular}
\end{sc}
\end{small}
\end{center}
\vskip -0.1in
\end{table}

In addition, to assess the detectability of SLS, we performed a Kolmogorov-Smirnov (K-S) test \citep{smirnov1948table} on the token rank distributions of greedy-decoded ($T=0$) and SLS-encoded sequences.

We observe that the CDF of the token ranks for encoded text closely tracks the original across all datasets, where the top 20 tokens hold over $95\%$ of the probability, indicating that SLS does not force the model into low-probability output trajectories.

The K-S metric reveals that the majority of samples across GSM8K, ShareGPT, and SWE-bench Verified yield a K-S statistic below 0.05. This distribution of K-S statistics confirms that the steganographic output is very difficult to distinguish statistically from the greedy generation (T=0) alone, making it robust to rank-based detection.

Specifically, in the distribution of K-S Statistics, the ShareGPT and GSM8K datasets show the highest levels of stealth, with the bulk of their distributions falling well to the left of the critical threshold. This confirms that the local entropy constraint effectively prevents encoding in sensitive, low-entropy regions where rank deviations would be most apparent.

\begin{figure}[h]
    \centering
    \includegraphics[width=\columnwidth]{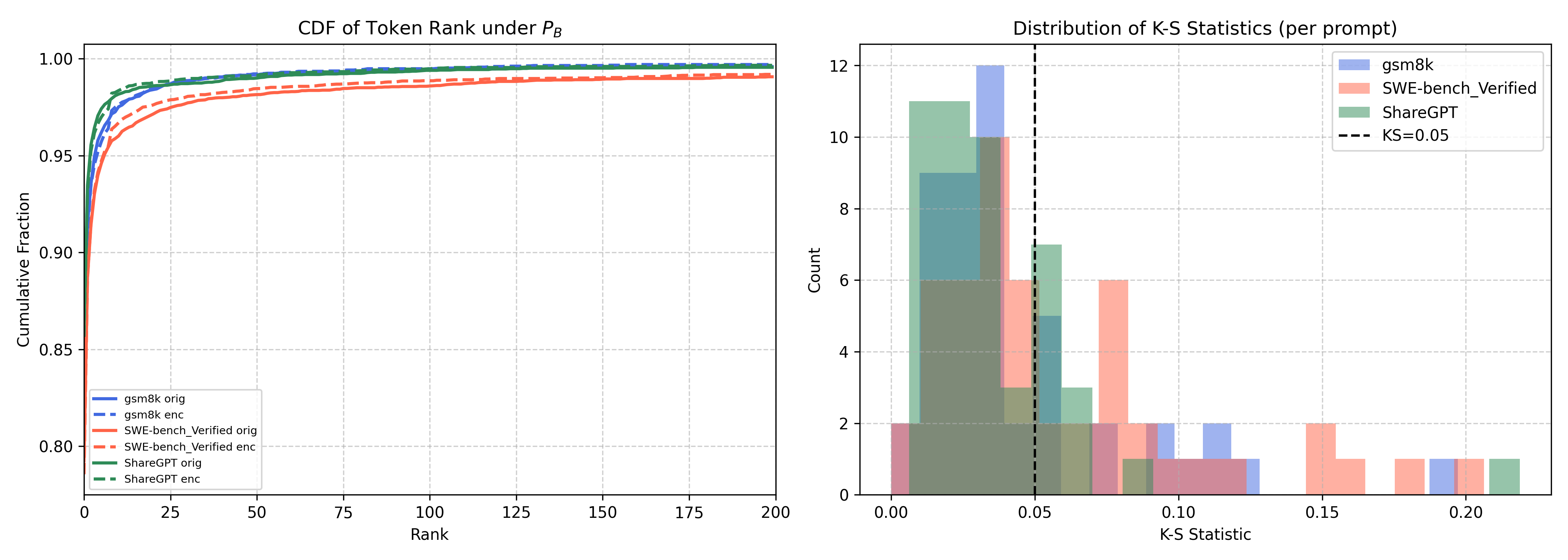}
    \caption{Kolmogorov-Smirnov test results comparing token rank distributions of greedy-decoded ($T=0$) and SLS-encoded sequences. The majority of prompts across all datasets yield a K-S statistic below 0.05, confirming that SLS outputs are statistically indistinguishable from normal greedy generation.}
    \label{fig:ks}
\end{figure}

Despite this statistical stealth, we observe a minor degradation in task-specific utility during high-capacity encoding. As shown in Table \ref{tab:payload_data}, the accuracy of \texttt{gemma-2-2b} on GSM8K decreases when payloads are embedded. This suggests a tradeoff between the preservation of complex reasoning chains and information density; even high-probability rank substitutions can steer the model away from the narrow manifold of correct logical tokens in technical domains. The tradeoff likely exists because of the threshold entropy being too high for these tasks. If the model embeds a payload in a numeric or logical token, the reasoning chain may get broken. Therefore, lowering the entropy threshold could ensure that the payload is encrypted in non-reasoning tokens. 

Consequently, deployment-ready implementations of SLS should incorporate an upstream task classifier to adaptively modulate encoding density. By restricting payload embedding to non-technical segments or conversational prompts, the system can preserve reasoning integrity while simultaneously increasing stealth; any detection heuristic applied to the excluded technical prompts would effectively be measuring null-signal noise, further obscuring the presence of the hidden channel.

Additionally, we evaluate the computational cost of each SLS variant. As shown in Table~\ref{tab:timing}, Single and Bursted encoding introduce negligible latency overhead ($+2.7\%$ and $+5.4\%$), while Periodic and Periodic Burst incur substantially higher cost (${\approx}88\%$) due to repeated dual forward passes over the full output sequence.

\begin{table}[h]
\centering
\caption{Timing overhead relative to normal greedy generation (A100, 10 runs, \texttt{gemma-2-2b}).}
\begin{tabular}{lcc}
\toprule
\textbf{Method} & \textbf{Avg. Time (s)} & \textbf{Overhead} \\
\midrule
Normal (Baseline) & 7.59 & --- \\
Single            & 7.79 & +2.7\% \\
Bursted           & 8.00 & +5.4\% \\
Periodic          & 14.32 & +88.7\% \\
Periodic Burst    & 14.24 & +87.7\% \\
\bottomrule
\end{tabular}
\label{tab:timing}
\end{table}

Finally, we evaluate end-to-end payload recovery accuracy across all four SLS variants on 20 ShareGPT prompts. SLS achieves 100\% decoding accuracy across all methods: 20/20 payloads for Single, 63/63 for Periodic, 58/58 for Burst, and 244/244 for Periodic Burst. For Burst encoding, 3 of the 20 prompts produced outputs where the local entropy never exceeded $\tau = 2.5$ within the burst window and were excluded from evaluation. These results confirm that the rank-based encoding is fully deterministic, assuming that the entropy constraint is satisfied.

\section*{Discussion}
This work introduces a deterministic steganographic method that enables covert communication through LLM-generated text without requiring a shared prompt context between sender and receiver. While the technique advances our understanding of how information can be embedded in language model outputs, it also raises dual-use concerns. On the beneficial side, SLS could inform legitimate applications such as watermarking model outputs, verifying content origin, or building defenses against hidden communication channels in deployed LLM systems. On the other hand, similar techniques could be misused to coordinate covert channels between AI agents, exfiltrate sensitive information from production LLM pipelines, or evade content moderation and oversight systems, risks that grow as LLMs are increasingly deployed in agentic and multi-agent settings.

We believe the benefits of publishing this work outweigh the risks, primarily because understanding how prompt-agnostic steganography can be constructed is a prerequisite for building effective defenses against it. Our results suggest several directions for mitigation, including monitoring for anomalous token-rank distributions, perturbing logits during decoding to disrupt rank-based encodings, and developing detectors trained on the statistical signatures of synchronized proxy prompts. We encourage future work on the detection, mitigation, and evaluation of such covert channels, particularly as LLM-based agents become more prevalent.

\section{Conclusion}
In this study, we introduce Synchronized Logit Steering (SLS), a simple, deterministic algorithm to encode payloads in situations where the sender does not know the intended receiver among a large number of recipients and cannot share internal vector databases or proprietary system prompts. By using the first $n$ output tokens as a synchronized prompt, we show that high-bit payloads can be hidden accurately by periodically encrypting multiple binary sequences. Overall, this work opens promising paths for real-time encryption and dynamic watermarking of LLM text generation. 

\bibliography{colm2026_conference}

@article{mathew_hidden_2024,
  title = {Hidden in {Plain} {Text}: {Emergence} \& {Mitigation} of {Steganographic} {Collusion} in {LLMs}},
  journal = {arXiv preprint arXiv:2402.07843},
  author = {Mathew, Yohan and Matthews, Ollie and McCarthy, Robert and Velja, Joan and Schroeder de Witt, Christian and Cope, Dylan and Schoots, Nandi},
  year = {2024}
}

@article{jiang2025high,
  title = {A high-capacity linguistic steganography based on entropy-driven rank-token mapping},
  author = {Jiang, Jun and Zhang, Weiming and Yu, Nenghai and Chen, Kejiang},
  journal = {arXiv preprint arXiv:2510.23035},
  year = {2025}
}

@inproceedings{ziegler2019neural,
  title = {Neural linguistic steganography},
  author = {Ziegler, Zachary and Deng, Yuntian and Rush, Alexander M.},
  booktitle = {Proceedings of the 2019 Conference on Empirical Methods in Natural Language Processing and the 9th International Joint Conference on Natural Language Processing (EMNLP-IJCNLP)},
  pages = {1210--1215},
  year = {2019}
}

@article{jaech2024openai,
  title = {Openai o1 system card},
  author = {Jaech, Aaron and Kalai, Adam and Lerer, Adam and Richardson, Adam and El-Kishky, Ahmed and Low, Aiden and Helyar, Alec and Madry, Aleksander and Beutel, Alex and Carney, Alex and others},
  journal = {arXiv preprint arXiv:2412.16720},
  year = {2024}
}

@article{lewis2020retrieval,
  title = {Retrieval-augmented generation for knowledge-intensive nlp tasks},
  author = {Lewis, Patrick and Perez, Ethan and Piktus, Aleksandra and Petroni, Fabio and Karpukhin, Vladimir and Goyal, Naman and K{\"u}ttler, Heinrich and Lewis, Mike and Yih, Wen-tau and Rockt{\"a}schel, Tim and others},
  journal = {Advances in neural information processing systems},
  volume = {33},
  pages = {9459--9474},
  year = {2020}
}

@inproceedings{ding2023discop,
  title = {Discop: Provably secure steganography in practice based on distribution copies},
  author = {Ding, Jinyang and Chen, Kejiang and Wang, Yaofei and Zhao, Na and Zhang, Weiming and Yu, Nenghai},
  booktitle = {2023 IEEE Symposium on Security and Privacy (SP)},
  pages = {2238--2255},
  year = {2023},
  organization = {IEEE}
}

@article{channalli_steganography_2009,
  author = {Channalli, Shashikala and Jadhav, Ajay},
  title = {Steganography {An} {Art} of {Hiding} {Data}},
  journal = {arXiv preprint arXiv:0912.2319},
  year = {2009}
}

@article{bieniasz_large_2024,
  title = {Large {Language} {Models} as {Carriers} of {Hidden} {Messages}},
  journal = {arXiv preprint arXiv:2402.17130},
  author = {Bieniasz, J\c{e}drzej and Janicki, Artur and Popio{\l}ek, Pawe{\l} and Hoscilowicz, Jakub},
  year = {2024}
}

@article{perry_robust_2025,
  author = {Perry, Neil and Pitta, Nishant and Rotem, Lior},
  title = {Robust {Steganography} from {Large} {Language} {Models}},
  journal = {arXiv preprint arXiv:2504.08977},
  year = {2025}
}

@article{bai_semantic_2024,
  author = {Bai, Minhao and Yang, Jinshuai and Pang, Kaiyi and Huang, Yongfeng and Gao, Yue},
  title = {Semantic {Steganography}: {A} {Framework} for {Robust} and {High}-{Capacity} {Information} {Hiding} {Using} {Large} {Language} {Models}},
  journal = {arXiv preprint arXiv:2412.11043},
  year = {2024}
}

@inproceedings{simmons_the_1984,
  author = {Simmons, Gustavus J.},
  title = {The {Prisoners'} {Problem} and the {Subliminal} {Channel}},
  booktitle = {Advances in {Cryptology}: {Proceedings} of {Crypto} 83},
  year = {1984}
}

@article{karpov_steganographic_2023,
  title = {The {Steganographic} {Potentials} of {Language} {Models}},
  journal = {arXiv preprint arXiv:2307.06925},
  author = {Karpov, Artem and Adeleke, Tinuade and Cho, Seong Hah and Perez-Campanero, Natalia},
  year = {2023}
}

@inproceedings{wu_generative_2024,
  title = {Generative {Text} {Steganography} with {Large} {Language} {Model}},
  booktitle = {Proceedings of the 32nd {ACM} {International} {Conference} on {Multimedia} ({MM} '24)},
  publisher = {ACM},
  author = {Wu, Jiaxuan and Wu, Zhengxian and Xue, Yiming and Wen, Juan and Peng, Wanli},
  year = {2024}
}

@inproceedings{li2023stegolm,
  title = {StegoLM: Linguistic Steganography via Pre-trained Language Models},
  author = {Li, Zichao and Peng, Wei and Liu, Xiang and Sun, Maosong},
  booktitle = {Proceedings of the 61st Annual Meeting of the Association for Computational Linguistics (ACL)},
  year = {2023}
}

@article{zhou2025autostega,
  title = {Auto-Stega: An Agent-Driven System for Lifelong Strategy Evolution in {LLM}-Based Text Steganography},
  author = {Zhou, Jiuan and Cheng, Yu and Xie, Yuan and Yin, Zhaoxia},
  journal = {arXiv preprint arXiv:2510.06565},
  year = {2025}
}

@article{raz2026canaries,
  title = {Safeguarding LLMs Against Misuse and AI-Driven Malware Using Steganographic Canaries},
  author = {Raz, Md and Putrevu, Venkata Sai Charan and Udeshi, Meet and Krishnamurthy, Prashanth and Khorrami, Farshad and Karri, Ramesh},
  journal = {arXiv preprint arXiv:2603.28655},
  year = {2026}
}

@article{kullback1951information,
  author = {Solomon Kullback and Richard A. Leibler},
  title = {On information and sufficiency},
  journal = {Annals of Mathematical Statistics},
  volume = {22},
  number = {1},
  pages = {79--86},
  year = {1951}
}

@article{smirnov1948table,
  author = {Smirnov, Nikolai V.},
  title = {Table for Estimating the Goodness of Fit of Empirical Distributions},
  journal = {Annals of Mathematical Statistics},
  volume = {19},
  number = {2},
  pages = {279--281},
  year = {1948}
}

@article{gemma2024,
  title = {Gemma 2: Improving Open Language Models at a Practical Size},
  author = {{Gemma Team}},
  journal = {arXiv preprint arXiv:2408.00118},
  year = {2024}
}

@article{cobbe2021gsm8k,
  title = {Training Verifiers to Solve Math Word Problems},
  author = {Cobbe, Karl and Kosaraju, Vineet and Bavarian, Mohammad and Chen, Mark and Jun, Heewoo and Kaiser, Lukasz and Plappert, Matthias and Tworek, Jerry and Hilton, Jacob and Nakano, Reiichiro and Hesse, Christopher and Schulman, John},
  journal = {arXiv preprint arXiv:2110.14168},
  year = {2021}
}

@article{jimenez2023swebench,
  title = {SWE-bench: Can Language Models Resolve Real-World GitHub Issues?},
  author = {Jimenez, Carlos E. and Yang, John and Wettig, Alexander and Yao, Shunyu and Pei, Kexin and Press, Ofir and Narasimhan, Karthik},
  journal = {arXiv preprint arXiv:2310.06770},
  year = {2023}
}
\bibliographystyle{colm2026_conference}

\end{document}